\documentclass[letterpaper, 10 pt, conference]{ieeeconf}  
\usepackage{amsmath,amsfonts}
\usepackage{algorithmic}
\usepackage{algorithm}
\usepackage{array}
\usepackage{graphicx}
\usepackage[caption=false, font=footnotesize]{subfig}
\usepackage{float}
\usepackage{textcomp}
\usepackage{stfloats}
\usepackage{url}
\usepackage{verbatim}
\usepackage{pifont}
\usepackage{scalerel}
\usepackage{cite}
\usepackage{soul}
\usepackage{multirow}
\usepackage{placeins}
\usepackage[dvipsnames]{xcolor}

\makeatletter
\newcommand*{\rom}[1]{\expandafter\@slowromancap\romannumeral #1@}
\newcommand*\bigcdot{\mathpalette\bigcdot@{.5}}
\newcommand*\bigcdot@[2]{\mathbin{\vcenter{\hbox{\scalebox{#2}{$\m@th#1\bullet$}}}}}
\makeatother

\IEEEoverridecommandlockouts

\begin{document}

\title{\LARGE \bf
Supervised Representation Learning towards Generalizable\\ Assembly State Recognition
}

\author{Tim J. Schoonbeek$^{1*}$, Goutham Balachandran$^{1}$, Hans Onvlee$^{2}$, Tim Houben$^{1}$, Shao-Hsuan Hung$^{1}$,\\ Jacek Kustra$^{2}$, Peter H.N. de With$^{1}$, Fons van der Sommen$^{1}$
\thanks{*Corresponding author, \tt\small t.j.schoonbeek@tue.nl.}
\thanks{$^{1}$Tim J. Schoonbeek, Goutham Balachandran, Tim Houben, Shao-Hsuan Hung, Peter H.N. de With, and Fons van der Sommen are with the Video Coding and Architectures research lab of the Department of Electrical Engineering, Eindhoven University of Technology, The Netherlands 
\tt\small \{t.houben,p.h.n.de.with,fvdsommen\}@tue.nl.}
\thanks{$^{2}$Hans Onvlee and Jacek Kustra are with ASML Research, The Netherlands 
\tt\small \{hans.onvlee, jacek.kustra\}@asml.com.}%
}

\maketitle

\begin{abstract}
Assembly state recognition facilitates the execution of assembly procedures, offering feedback to enhance efficiency and minimize errors. However, recognizing assembly states poses challenges in scalability, since parts are frequently updated, and the robustness to execution errors remains underexplored. To address these challenges, this paper proposes an approach based on representation learning and the novel intermediate-state informed loss function modification (ISIL). ISIL leverages unlabeled transitions between states and demonstrates significant improvements in clustering and classification performance for all tested architectures and losses. Despite being trained exclusively on images without execution errors, thorough analysis on error states demonstrates that our approach accurately distinguishes between correct states and states with various types of execution errors. The integration of the proposed algorithm can offer meaningful assistance to workers and mitigate unexpected losses due to procedural mishaps in industrial settings. The code is available at: \\
   {\tt\small \url{https://timschoonbeek.github.io/state_rec}}
\end{abstract}


\section{Introduction}
\label{sec:intro}
Correctly recognizing the assembly state of a particular object during a procedural action can ensure that the procedure is being executed properly and allow for dynamic follow-up instructions on the fly. This assembly state recognition is especially relevant in industrial settings, where the objects under consideration are usually complex and prone to execution errors during assembly. Such errors can invoke intensive maintenance procedures and loss of material. Assembly state recognition can help prevent errors and optimize the overall efficiency of procedures, for instance, via integration into augmented reality devices such as the HoloLens~\cite{su2019deep} (Microsoft, Seattle, USA).

\begin{figure}
    \centering
    \includegraphics[width=0.99\columnwidth, trim={0 3cm 0 0}]{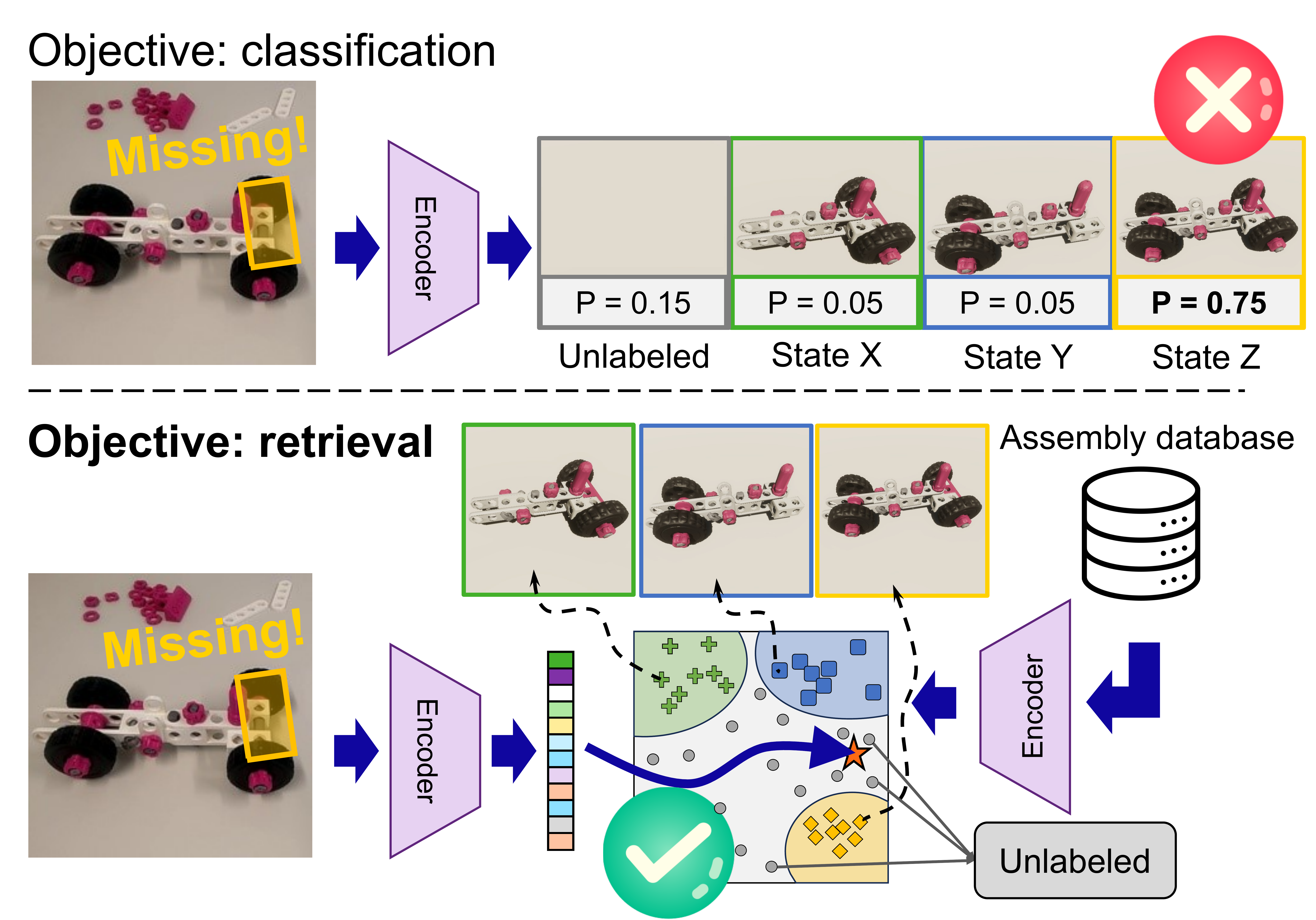}
    \caption{This work defines assembly state recognition as a representation learning approach (bottom of figure), rather than classification (top), and demonstrates that this (1) outperforms classification-based approaches, (2) avoids re-training the models after every minor update in the assembly procedure, and (3) enables distinction between correct and erroneous states.}
    \label{fig:teaser}
\end{figure}

Existing methods towards assembly state recognition or detection are approaching the problem from a classification perspective, training a neural network to memorize the expected object states~\cite{stanescu2023state,su2019deep,liu2020tga,zhou2020fine,pang2020marker,murray2024equipment}. 
Although this has shown decent performance for procedures with a few expected assembly states, such approaches require the explicit definition of each possible assembly state~\cite{jones2021fine}. 
Therefore, if any of the object states are modified, \textit{e.g.} due to an upgraded part or a newly added state, new data must be annotated and the entire network should be re-trained or fine-tuned. 
This is a blocking issue, and apart from being costly it significantly hinders scalability to industrial applications. 
A second concern is the under-investigation of the performance of existing approaches on error states, where the approaches remain incapable of reliably detecting erroneous (or correct, but previously undefined) assembly states that are not explicitly defined in training. Requiring explicit definitions of error states is a fundamental issue, since defining all possible errors that can occur during an assembly is infeasible (and unexpected and thus undefined errors are frequently the most troublesome). We propose to use a representation learning framework to address these challenges by learning a function that projects images into a meaningful embedding space.

By using representation learning, rather than explicitly memorizing each possible assembly state, the model is able to distinguish between states based on visual similarity. We hypothesize that such an approach is able to generalize to new assembly states and (unseen) execution errors, as it highly resembles open-set classification and image retrieval tasks~\cite{musgrave2020metric}. To test this hypothesis, we present a representation learning framework for assembly state recognition. Furthermore, we propose a simple yet effective modification to existing loss functions, named intermediate-state informed loss (ISIL), which uses unlabeled part configurations in the training data as negative samples without enforcing their image embeddings towards a single cluster. The improvements obtained by using ISIL on clustering performance range from 5\% to 22\%, across various network architectures and contrastive loss functions, whilst also improving classification performance on all experiments. We show that our framework, trained on a combination of real-world and synthetic images, outperforms classification-based approaches and generalizes to entirely unseen assembly states. Furthermore, we demonstrate that representation learning-based approaches recognize real-world assembly errors from practical cases better than classification-based approaches, without being trained on any data containing error states. The annotations created to perform the study on these error states are published to stimulate research on this topic. This work contains the following contributions:
\begin{itemize}
    \item \textit{Representation learning approach} to assembly state recognition that outperforms classification approaches and performs well even on unseen part configurations;
    \item \textit{Intermediate-state informed loss function modification} ISIL, which effectively leverages unlabeled data between completed assembly steps to improve both clustering and classification performance;
    \item \textit{Extensive study on erroneous assembly states} for the representation learning framework, compared to classification, including new annotations for intended states in the IndustReal~\cite{schoonbeek2024industreal} dataset.
\end{itemize}

\section{Related Work}
\label{sec:relwork}

\subsection{Assembly State Detection}
\label{sec:asd}
Assembly state detection (ASD) is generally approached as a subset of the broader object detection problem~\cite{schoonbeek2024industreal,zhou2020fine,murray2024equipment,schieber2024asdf}. ASD differs from assembly state recognition, as there is no set of reference images, and the objective is to classify and localize different states of the assembly process. 
In an assembly process, there may be many states that are highly similar and therefore hard to distinguish individually. This issue is addressed by Stanescu~\textit{et al.}~\cite{stanescu2023state} using a modified object detector to include previous state detections, and in the work by Liu~\textit{et al.}~\cite{liu2020tga} with specialized attention units. 
Another work uses a neural network trained to simultaneously perform pose estimation and state classification~\cite{su2019deep}. 

Despite the significant progress made on ASD, current approaches have two notable limitations. First, they focus on classification with a fixed set of assembly states, since the problem is addressed as conventional object detection. Therefore, even if the models perform well on these pre-defined states, full re-training is required for any changes in the assembly procedures. 
One related work does not rely on classification, however, they require multiple depth sensors arranged in a fixed set-up and a restrictive step-by-step assembly execution~\cite{pang2021image}. 
Second, although it is common to encounter unexpected states in complex assembly tasks (\textit{e.g.} due to execution errors), and this is studied in robot assembly tasks~\cite{elhafsi2023semantic}, existing research on state recognition fails to investigate this~\cite{su2019deep,zhou2020fine,schieber2024asdf}.
None of the existing approaches generalize to new, unseen error states. We address both of the aforementioned limitations by approaching assembly state recognition as a representation learning task.

\begin{figure*}
  \centering
  \subfloat[\textbf{Exclude intermediate states} from training, focusing exclusively on pre-defined classes.\label{fig:no_inter}]{%
       \includegraphics[width=0.3\linewidth, trim={0 0 0 0.5cm}]{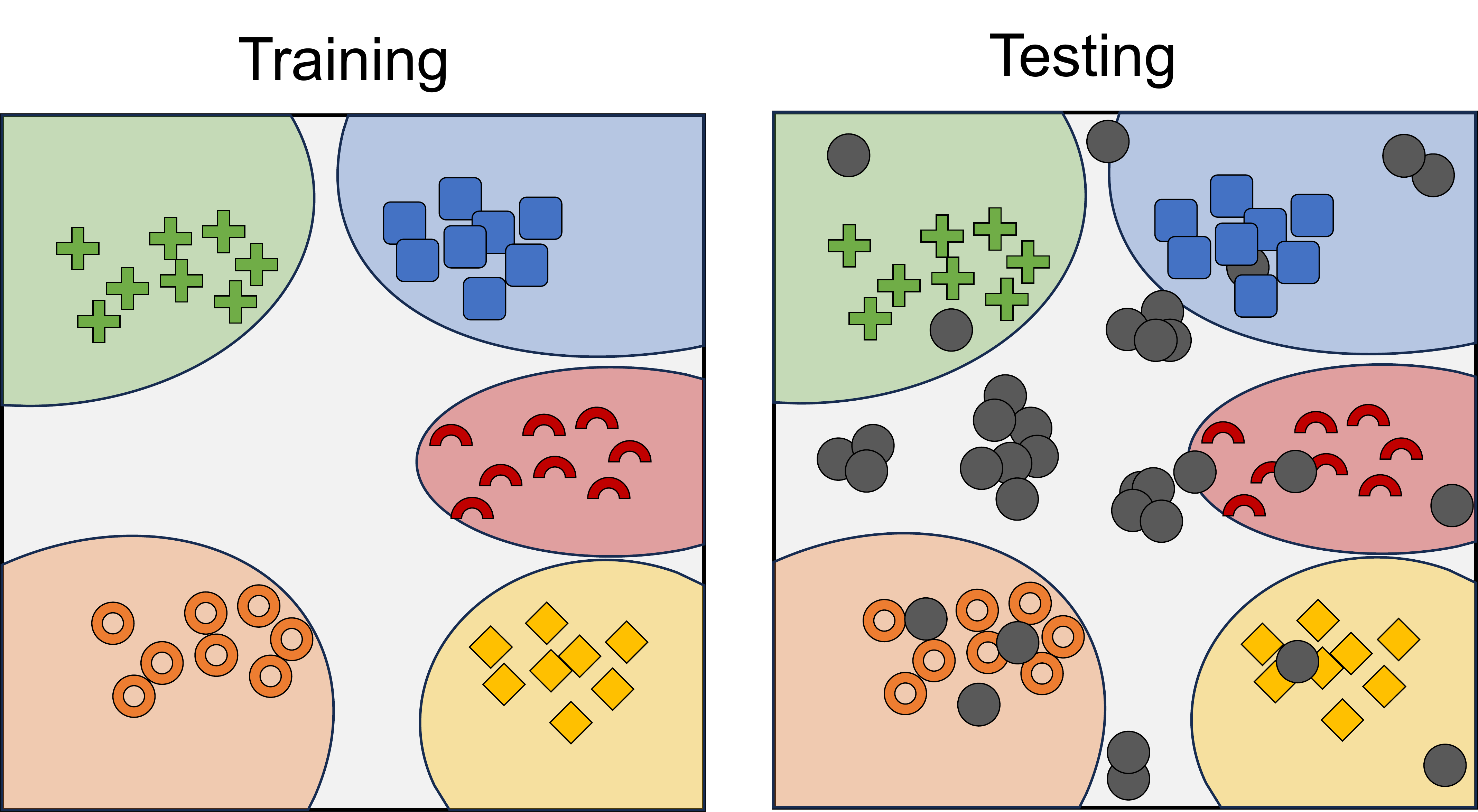}}
    \hfill
  \subfloat[\textbf{Include intermediate states} without modifications, treating them as a separate class.\label{fig:inter}]{%
        \includegraphics[width=0.3\linewidth, trim={0 0 0 0.5cm}]{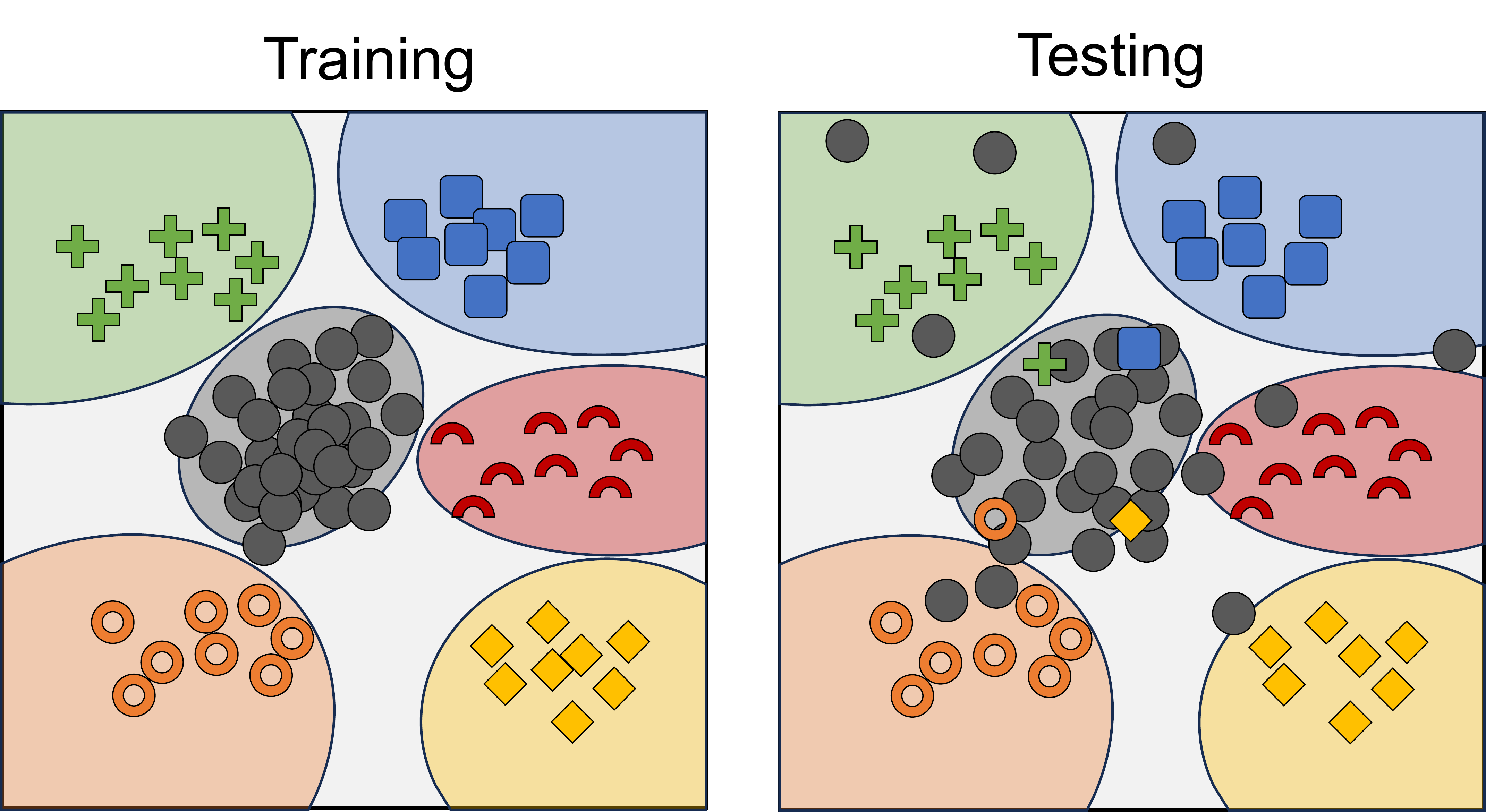}}
    \hfill
  \subfloat[\textbf{Intermediate-state informed loss}: exclusively use the undefined samples as negatives.\label{fig:inter_ex}]{%
        \includegraphics[width=0.3\linewidth, trim={0 0 0 0.5cm}]{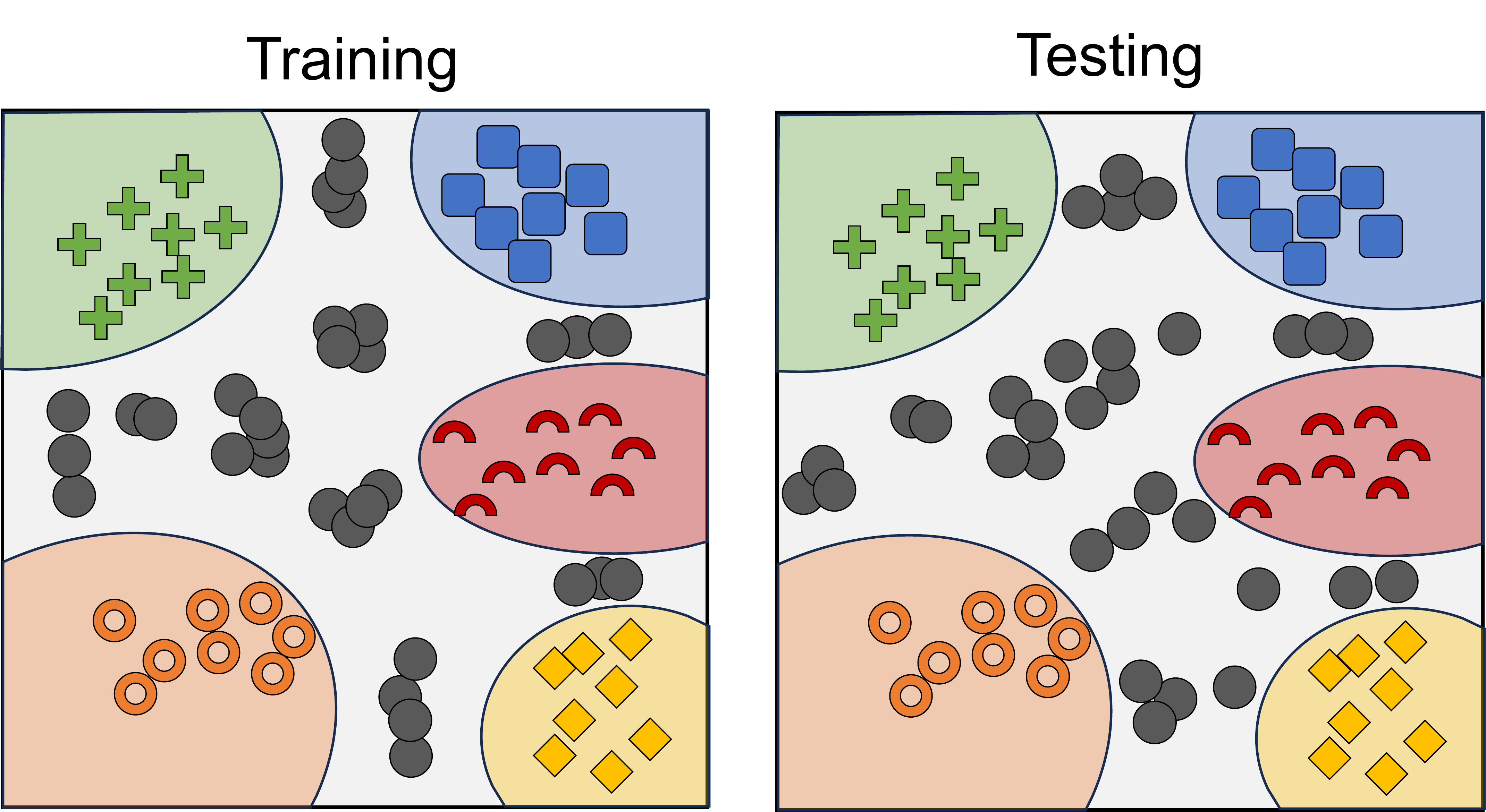}}
    \hfill
  \caption{Illustration of influence of the proposed loss function that leverages images containing intermediate assembly states, \textit{i.e.} non-defined, transitional states between pre-defined states.
  In (a), the intermediate states (gray) are ignored during training, not leveraging any information that these images might have. 
  In (b), the intermediate states are grouped into a single cluster, hindering the model's capacity to capture a meaningful embedding since these states are frequently not correlated. 
  We propose (c), an intuitive modification to loss functions that leverages intermediate states exclusively as negative samples. 
  Dissimilar embeddings of (potentially) uncorrelated states are only penalized if they are similar to any foreground (pre-defined) class.}
  \label{fig:method}
  \vspace{-0.33cm}
\end{figure*}

\subsection{Representation Learning}
\label{sec:rep_learn}
In representation learning, the objective is to learn a function that projects an input image to a meaningful embedding space. 
One way to implement such a function is contrastive learning, where the goal is to learn a representation of the data such that similar instances are in close proximity in feature space, while dissimilar instances are far apart~\cite{simclr,simclrv2,dino}. This can be achieved by, \textit{e.g.} maximizing agreement between differently augmented views of the same data example via a contrastive loss in the latent space~\cite{simclr,simclrv2}, or with Siamese neural networks (SNNs)~\cite{snn1,oneshotsiamese}. SNNs consist of two encoders with shared weights and have been applied to a variety of verification tasks, such as signature~\cite{snn1} and face verification~\cite{snn2}. During training, the model extracts features from two images, after which a distance metric between these features determines the similarity between them. 

A commonly used loss function in representation learning is the triplet loss~\cite{facenet}, with the objective to ensure that the projected point of an anchor sample is closer to the projected point of a positive than to a negative sample. The number of possible triplets grows cubically with the size of the dataset and a large part of all triplets will be trivial, introducing the need for hard mining. To that end, Hermans~\textit{et al.}~\cite{triplet} propose the Batch Hard triplet loss, where the hardest positive and negative for each image within a mini-batch is calculated using the Euclidean distance. The InfoNCE~\cite{infonce} loss is another commonly used loss function, popular in unsupervised learning frameworks such as SimCLR~\cite{simclr,simclrv2}, and aims to maximize mutual information between positive samples whilst minimizing information between negatives. The supervised contrastive (SupCon)~\cite{supcon} loss function extends the InfoNCE loss to a supervised setting. 

A significant advantage of representation learning is its ability to generalize~\cite{oneshotsiamese,snn2}. Whereas classification and detection networks are trained to discriminate samples among previously seen/memorized categories, representation learning networks learn a correspondence task. Therefore, the model does not explicitly learn to memorize the training data, but to learn low-dimensional, discriminative features. These features can be used not only to generalize to new instances of known classes, but to entirely new classes from unknown distributions (one-shot learning)~\cite{oneshotsiamese}. Hence, we hypothesize that the network can generalize to unseen part configurations, as well as error states. This makes the representation learning approach a viable and attractive candidate for solving the challenges encountered in assembly state recognition.

\section{Method}
\label{sec:method}
The proposed representation learning approach, as outlined in Fig.~\ref{fig:teaser}, aims to extract meaningful embeddings from images of various assembly stages. At inference, the embeddings of assembly images are used as query and compared to the training embeddings, in order to recognize the state (and thereby the correctness) of assemblies. Additionally, we describe a loss function modification that can be implemented to any supervised contrastive loss function.

\subsection{Contrastive Learning Framework}
\label{sec:con_framework}
For the contrastive learning framework, the general structure proposed in SimCLRv2~\cite{simclrv2} is selected and used in a supervised setting. Instead of creating multiple instances of one image by applying various augmentations, $M_R$ real-world and $M_S$ synthetic images are sampled for $N$ randomly selected assembly states per batch. Additionally, $I_R$ images of unlabeled assembly images are sampled, resulting in a batch size of $(M_R+M_S)\cdot N + I_R$. These unlabeled images can contain transitions between assembly states, as further outlined in Sec.\ref{sec:ISIL}, but can also contain images where the assembly is largely or entirely out-of-frame. 
No explicit domain adaptation techniques beyond standard image augmentations are employed. The synthetic images are generated with a simulator~\cite{borkman2021unity}, using the CAD models for different assembly stages. The use of synthetic images provides more samples for each assembly state during training, which is particularly meaningful for the recognition of assembly states rarely present in real-world training data.

The objective during training is to learn a function that maps identical physical assembly states (with different orientations and backgrounds) from the input manifold $\mathcal{R}^F$ onto close points in the embedding space $\mathcal{R}^D$, while simultaneously mapping different assembly states to distant points in $\mathcal{R}^D$. This mapping function is learned by passing images through an encoder $f(\cdot)$ that creates feature maps $h_i$. The encoder can be any image feature extractor, such as a ResNet~\cite{he2016deep} or vision transformer (ViT)~\cite{dosovitskiy2020image}. These feature maps are then projected to the embeddings $z\in\mathcal{R}^D$ using projection head $g(\cdot)$, implemented as a three-layer multi-layer-perceptron (MLP), after which the features are normalized. Adding a projection head and defining the contrastive loss in $\mathcal{R}^D$, rather than directly using the feature maps $h_i$, leads to improved representations~\cite{simclr}. 
An overview of the framework is described in Fig.~\ref{fig:method}.

At the inference stage, images are passed through the encoder $f(\cdot)$. Given~\cite{simclrv2}, we discard the last two layers of the three-layer MLP $g(\cdot)$ to create the final embeddings $y_i$. This is useful because using a middle layer of the projection head $g(\cdot)$ improves the performance compared to discarding the entire head. To determine the assembly state in a test image, its embedding $\hat{y_i}$ is compared to all reference embeddings~$y$ from the training data using K-Nearest Neighbour (KNN). Neighbours are determined with same similarity metric as used in the loss function during training, \textit{i.e.} Euclidean distance ($L_2$) for the triplet loss and cosine similarity for models trained with an angular-based loss function.

\begin{figure*}
    \centering
    \vspace{5pt}
    \includegraphics[width=1.75\columnwidth, trim={0 0cm 0 0}]{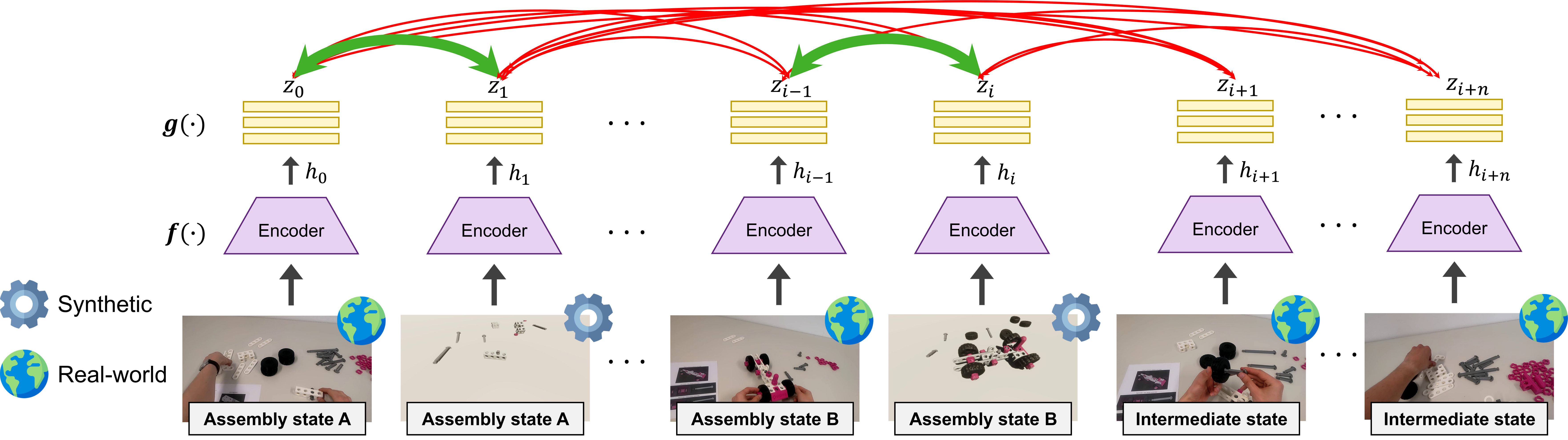}
    \caption{Overview of the contrastive learning framework (with the proposed ISIL modification). Each mini-batch consists of real-world and synthetic images of pre-defined assembly states and real-world intermediate states. The images are passed through an encoder $f(\cdot)$, followed by a three-layer MLP projection head $g(\cdot)$. The resulting embeddings $z_i$ are used to calculate the contrastive loss. During inference, only the first layer of $g(\cdot)$ is used.} 
    \label{fig:method}
    \vspace{-0.33cm}
\end{figure*}

\subsection{Intermediate-State Informed Loss}
\label{sec:ISIL}
During the execution of an assembly or maintenance procedure, an object goes through various assembly states (i.e., part configurations). For instance, in the procedure of attaching a wheel with 5 log nuts to a vehicle, there are $2^5=32$ potential part configurations. Defining all potential states is therefore frequently infeasible, resulting in \textit{undefined} transitional states between explicitly defined states (milestones) in videos of such procedures. We define these transitional states as \textit{intermediate states}. Discarding data containing these states during training is sub-optimal, as the configurations being undefined, does not mean that there cannot be information within these configurations, as illustrated in Fig.~\ref{fig:no_inter}. On the other hand, since the undefined states can contain any possible transition between any possible assembly state, these intermediate states are frequently not correlated to each other. Therefore, simply classifying or clustering these intermediate states into a single group forces a neural network to learn non-existing connections, which may limit the embedding capacity of a model.

To overcome this limitation, this work proposes an intuitive modification that can be applied to supervised contrastive loss functions. The proposed modification, \textit{intermediate-state informed loss} (ISIL), leverages the unlabeled intermediate assembly states by treating them exclusively as (hard) negative samples. By doing so, intermediate states are forced to be dissimilar to any explicitly defined assembly state, whilst not penalizing dissimilar embeddings of intermediate states, or enforcing similarity between embeddings of potentially uncorrelated training samples. This concept is illustrated in Fig.~\ref{fig:inter_ex}. 

The ISIL modification can be applied to different loss functions for representation learning. In this work, we apply it to two common losses, namely the Batch Hard (triplet)~\cite{triplet} and supervised contrastive (SupCon) loss~\cite{supcon}. The former is defined, with the ISIL modification underlined, as 
\begin{align}
    L_{\text{triplet}} = \sum_{i=1}^{N} \bigg[m&+\overbrace{\max_{\substack{j: y_j = y_i \underline{\land y_i\in C}}} \| g(f(x_i)) - g(f(x_j)) \|_2^2}^{\text{hardest positive}} \nonumber \\
    & - \underbrace{\min_{\substack{k: y_k \neq y_i}} \| g(f(x_i)) - g(f(x_k)) \|_2^2}_{\text{hardest negative}} \bigg]_+ 
\label{eq:t1}
\end{align}
where $f(x)$ is the embedding function described in Sec.~\ref{sec:con_framework}, $C$ the set of pre-defined assembly classes, $N$ the number of samples in the mini-batch, $m$ the margin, $x_j$ a positive and $x_k$ a negative for the given anchor input $x_i$, with $y_j$, $y_k$, and $y_i$ as their respective labels. The ISIL modification is enforced by ${y_i\in C}$, denoting that the ground-truth label of sample $i$ is a pre-defined assembly state (therefore not an intermediate state). This excludes undefined states from the \textit{pulling} contribution within the triplet loss, ensuring that undefined samples are not grouped towards one cluster. 

Similarly, the SupCon~\cite{supcon} loss function with the ISIL modification is defined as
\begin{equation}
    \mathcal{L}_{\text{SupCon}} = \sum_{i\in I} \log \frac{-1}{\lvert \underline{P(i)}\rvert} \sum_{p\in \underline{P(i)}} \frac{\exp(\frac{g(f(x_i)) \cdot g(f(x_p))}{\tau})}{\sum \limits_{a\in A(i)} \exp(\frac{g(f(x_i)) \cdot g(f(x_a))}{\tau})},
    \label{eq:t2}
\end{equation}
where $\tau$ denotes the temperature constant, ${A(i) \equiv X \setminus \{i\}}$ the set of all indices in mini-batch $X$, and ${\underline{P(i)}\equiv \{ p \in A(i) : (y_p = y_i) \land \underline{(y_p\in C)}\}}$ is the set of indices of all positives for defined classes. The ISIL modification is enforced by excluding the undefined states from the set of positives indices $P(i)$, while permitting them to act as negatives in $A(i)$.

\section{Experiments}
\label{sec:experiments}

The representation framework as described in Section \ref{sec:con_framework} is implemented, as well as the intermediate-state informed loss function modification (ISIL) for the batch hard (triplet)~\cite{triplet} and SupCon~\cite{supcon} losses. To support our claim that representation learning is effective for assembly state recognition, we compare our method to a classification model, trained with the cross-entropy loss, where the last layer of the projection head is modified to equal the number of defined assembly states in the training data. At inference, only the first layer of projection head $g(\cdot)$ is retained, regardless of the loss used in training, which means no further modifications are required to directly compare the embeddings of a classification-trained model to those from a contrastive-trained model.

To evaluate the use of representation learning for assembly state recognition, the IndustReal~\cite{schoonbeek2024industreal} dataset is used. This dataset selected for three reasons. Firstly, the data contain 22~annotated assembly states, making up only 13\% of the total images. Therefore, significant information is discarded by training only on pre-defined states. Secondly, all parts in the IndustReal assembly are provided as 3D models. This facilitates the creation of data with entirely new part configurations, which we use to compare the generalization of our framework to the model trained for classification. Finally, IndustReal contains various execution errors, which we use to the test our hypothesis that the generalization capacity of representation learning does not only benefit the recognition of novel part configurations, but also enables the verification of the correctness of assembly states. Models are trained following the IndustReal train/val/test splits~\cite{schoonbeek2024industreal}, with the exception that we exclude erroneous assembly states during training. Therefore, the model's generalization to entirely unseen errors can be tested.

In order to quantify the performance of a trained model, we report the $F_1\text{@}1$ and $\textit{MAP}\text{@}R$ scores. The $F_1\text{@}1$ metric represents the harmonic mean of precision and recall at the top of the ranked list of closest neighbours. The mean average precision at $R$ ($\textit{MAP}\text{@}R$) score is a metric used to evaluate the quality of the ranked retrieval results for a given query~\cite{musgrave2020metric}. It calculates the mean of the average precision scores at $R$, where $R$ is the number of true positives in the reference set, thereby reflecting the precision of the system across different recall levels for the top $R$ results. $\textit{MAP}\text{@}R$ is more stable than Recall@$1$ and takes the order of the ranking of correct retrievals into account, unlike $R$-Precision. As mentioned above, intermediate states are frequently not correlated to each other. Since the embeddings of two dissimilar intermediate states are allowed (and indeed desired to be) dissimilar to each other, it does not make sense to evaluate $\textit{MAP}\text{@}R$ on these classes. Therefore, we report $\textit{MAP}\text{@}R$ on the defined assembly states only and denote this with $\textit{MAP}\text{@}R(+)$. Please note that this score is still strongly impacted by intermediate states, as these states should not appear within the clusters of defined assembly states.

\subsection{Implementation Details}
The proposed representation learning approach consists of an image encoder and projection head, as outlined in Section \ref{sec:method}. For the encoders, ImageNet-1k pre-trained ResNet34~\cite{he2016deep} and ViT-S~\cite{vit} are implemented and compared to each other. The last fully connected layer of the encoders is replaced with three fully connected layers to form the projection head. The first two layers maintain the feature embedding size of the encoder, whilst the last layer reduces the features onto a 128-dimensional embedding space. 

Each image is resized to 224$\times$224 pixels ($W\times H$). For each batch, real-world ($M_R$=8) and synthetic images ($M_S$=8) are sampled per assembly state for $N$=15 randomly selected states. Additionally, $I_R$=64 intermediate assembly states are sampled for the loss functions that do not discard these images, resulting in a batch size of 240 or 304, depending on the use of intermediate states.
The margin ($m$=0.01) and temperature ($\tau$=0.07) values for the loss functions are empirically set using the validation set.

\begin{figure}
  \centering
  \vspace{5pt}
  \subfloat{%
       \includegraphics[width=0.99\columnwidth, trim={0 0 0 0}]{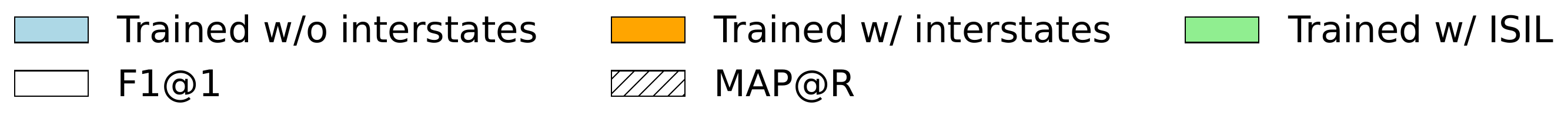}}
    \hfill
  \setcounter{subfigure}{0}
  \subfloat[ResNet-34~\cite{he2016deep} backbone.\label{fig:res_nointer_industreal}]{%
        \includegraphics[width=0.485\columnwidth, trim={0 0 0 1.5cm}]{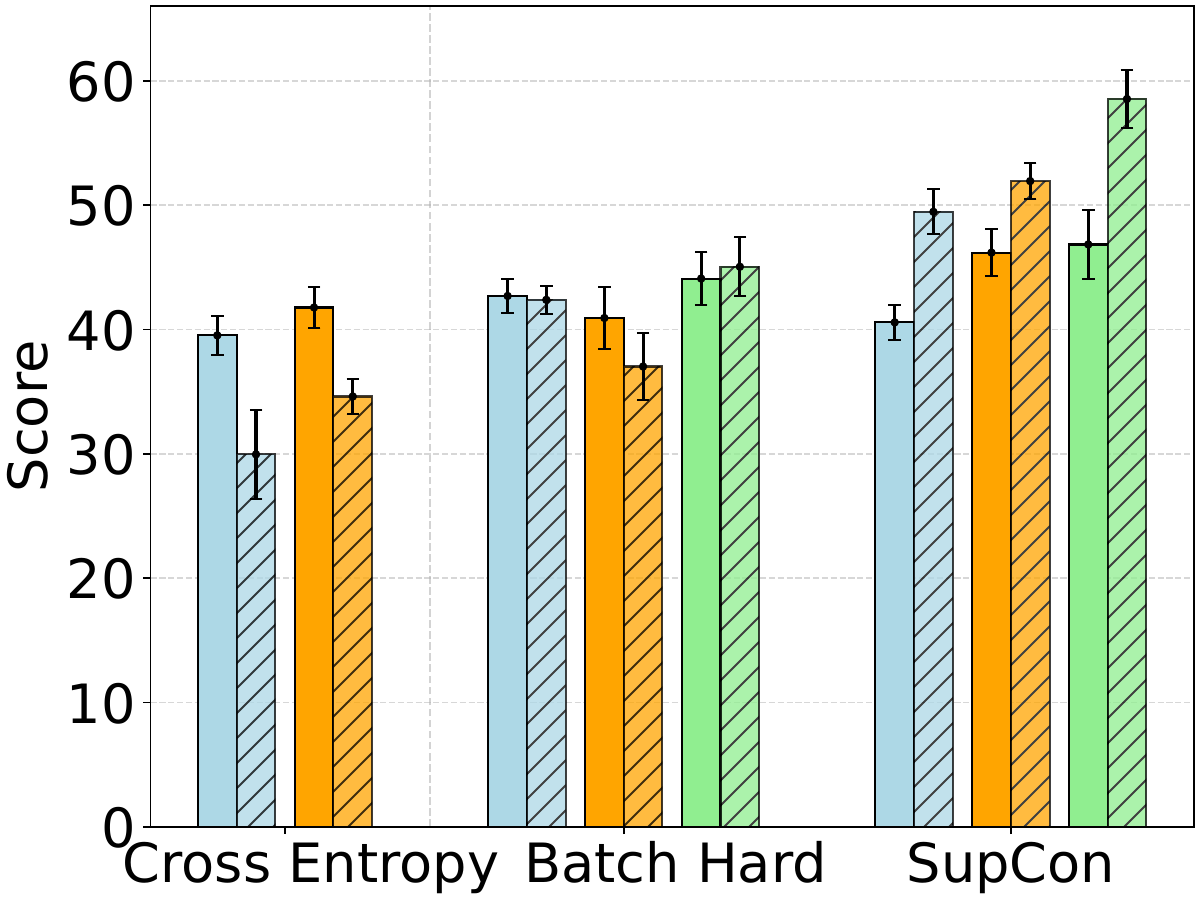}}
    \hfill
  \subfloat[ViT-S~\cite{vit} encoder.\label{fig:res_inter_industreal}]{%
        \includegraphics[width=0.485\columnwidth, trim={0 0 0 1.5cm}]{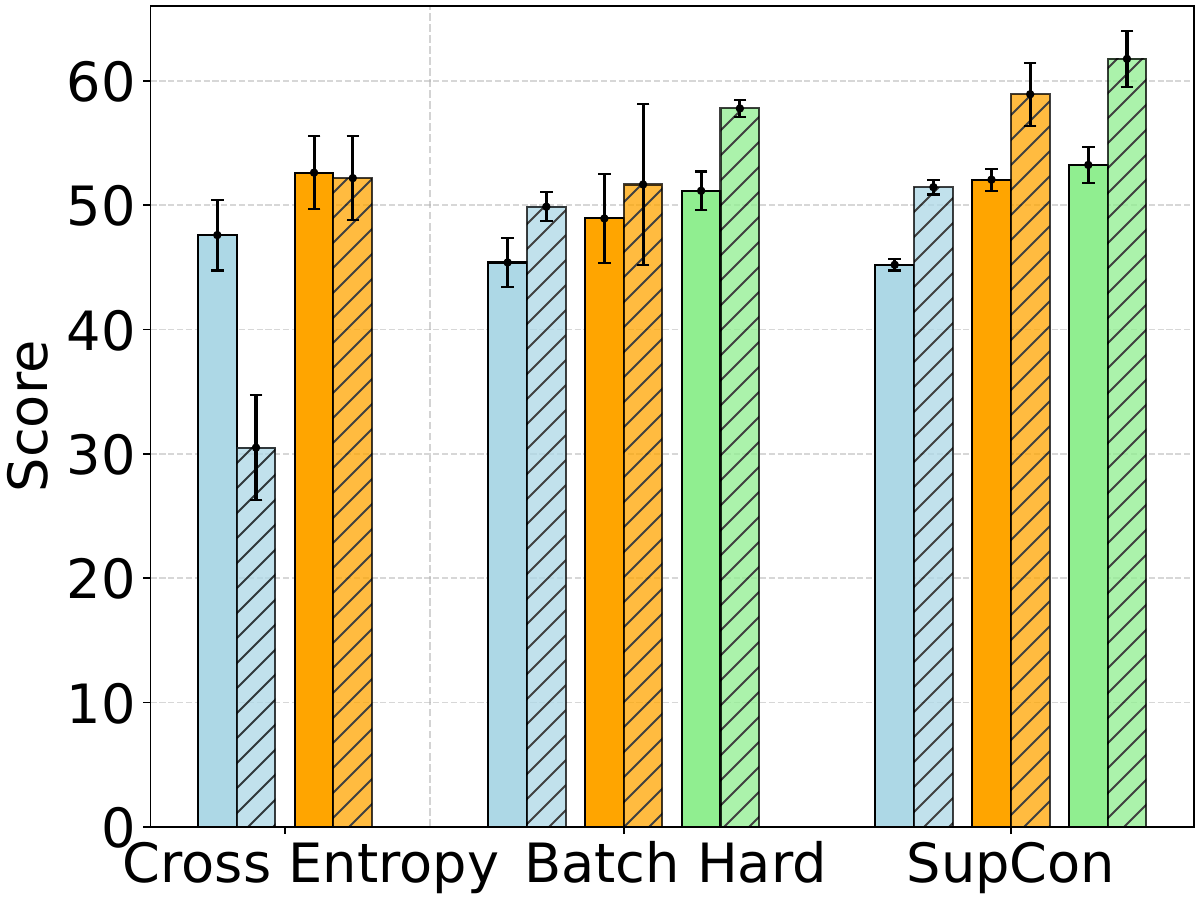}}
    \hfill
  \caption{Assembly state recognition performance on IndustReal~\cite{schoonbeek2024industreal}. The contrastive approaches outperform those trained for classification, and the proposed ISIL modification increases performance in all settings.}
  \label{fig:industreal_test}
  \vspace{-0.45cm}
\end{figure}

For consistency between training cycles, the Adam optimizer with a learning rate and weight decay of 10$^{-4}$ are selected. A 15k iteration (linear) warm-up learning rate is employed, followed by a cosine annealing learning rate with warm restarts every 40k iterations~\cite{loshchilov2016sgdr}. The best performing models are selected based on $F_1\text{@}1$ on the IndustReal validation set. After 100k iterations, or once a model does not improve either in $F_1\text{@}1$ or $\textit{MAP}\text{@}R(+)$ on this validation set for 20k iterations, training is stopped. For every experiment configuration, five cycles with different seeds are performed. Whilst the backbones are pre-trained and thus constant between cycles, the projection layer weights and batches are generated based on the seed, and therefore different per cycle. For data augmentation during training, random Gaussian blur (kernel size 5, $\sigma$=2) and brightness, saturation, and contrast jitter of at most 0.1, 0.7, and 0.1 respectively, are applied. The framework executes in real time on a Tesla~v100 GPU, obtaining approximately 150~frames/sec~(fps) per image with either backbone.

\subsection{Assembly State Recognition on IndustReal}
\label{sec:exp_staterecognition}
Figure~\ref{fig:industreal_test} demonstrates the results on the IndustReal~\cite{schoonbeek2024industreal} test set for the ResNet-34 and ViT-S backbones, trained with the cross-entropy, batch hard, and SupCon loss functions. 
The test set contains 9,659 images of assembly states, with 537$\pm$413 images per state for 18 states. Furthermore, the test set contains 20,101 images containing intermediate states or background, 25\% of the total unlabeled images in the IndustReal test set to significantly improve computational load. The reduction of unlabeled images is achieved by taking every fourth image with equal time spacing between the discarded images.
Given the inherent class imbalance between assembly states in IndustReal, we report macro-averaged metrics. On average, the best contrastively-trained ResNet model outperforms the classification-trained ResNet by 12\% on $F_1\text{@}1$, a notable result given that the metric evaluates classification rather than clustering performance. On $\textit{MAP}\text{@}R(+)$, the SupCon loss improves over the cross-entropy loss by 69\%. In fact, for the ResNet encoder, all configurations of the contrastive losses outperform the cross-entropy loss, regardless of whether any intermediate states are used during training. The ViT encoder trained with cross-entropy performs significantly better than ResNet trained for classification, but still scores 16\% lower on $\textit{MAP}\text{@}R(+)$, compared to the best contrastive-trained ViT. These results clearly demonstrate the benefit of representation learning for assembly state recognition.

\begin{figure}
  \centering
  \vspace{5pt}
  \subfloat{%
       \includegraphics[width=0.99\columnwidth, trim={0 0 0 0}]{legend_industreal.pdf}}
    \hfill
  \setcounter{subfigure}{0}
  \subfloat[ResNet-34~\cite{he2016deep} backbone.\label{fig:res_nointer_gen}]{%
        \includegraphics[width=0.485\columnwidth, trim={0 0 0 1.5cm}]{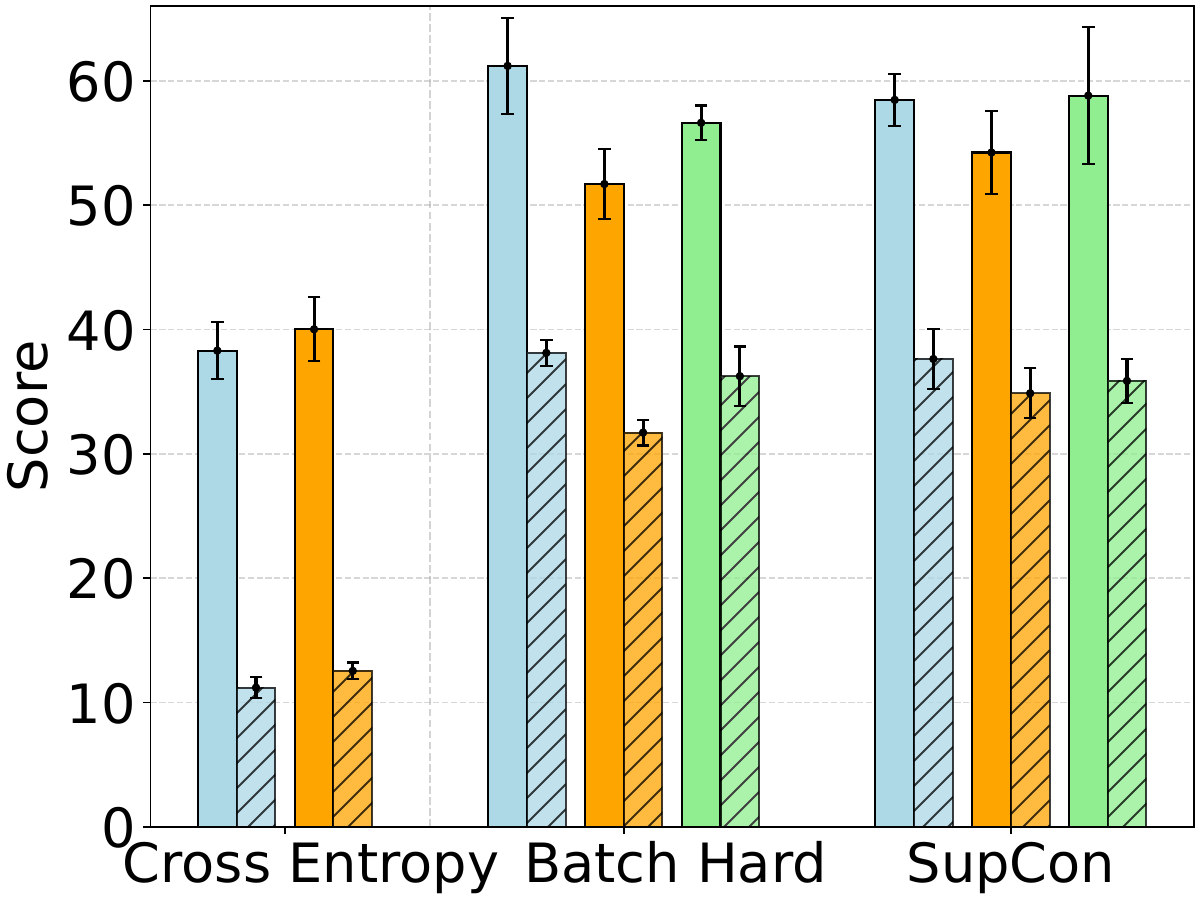}}
    \hfill
  \subfloat[ViT-S~\cite{vit} encoder.\label{fig:res_inter_gen}]{%
        \includegraphics[width=0.485\columnwidth, trim={0 0 0 1.5cm}]{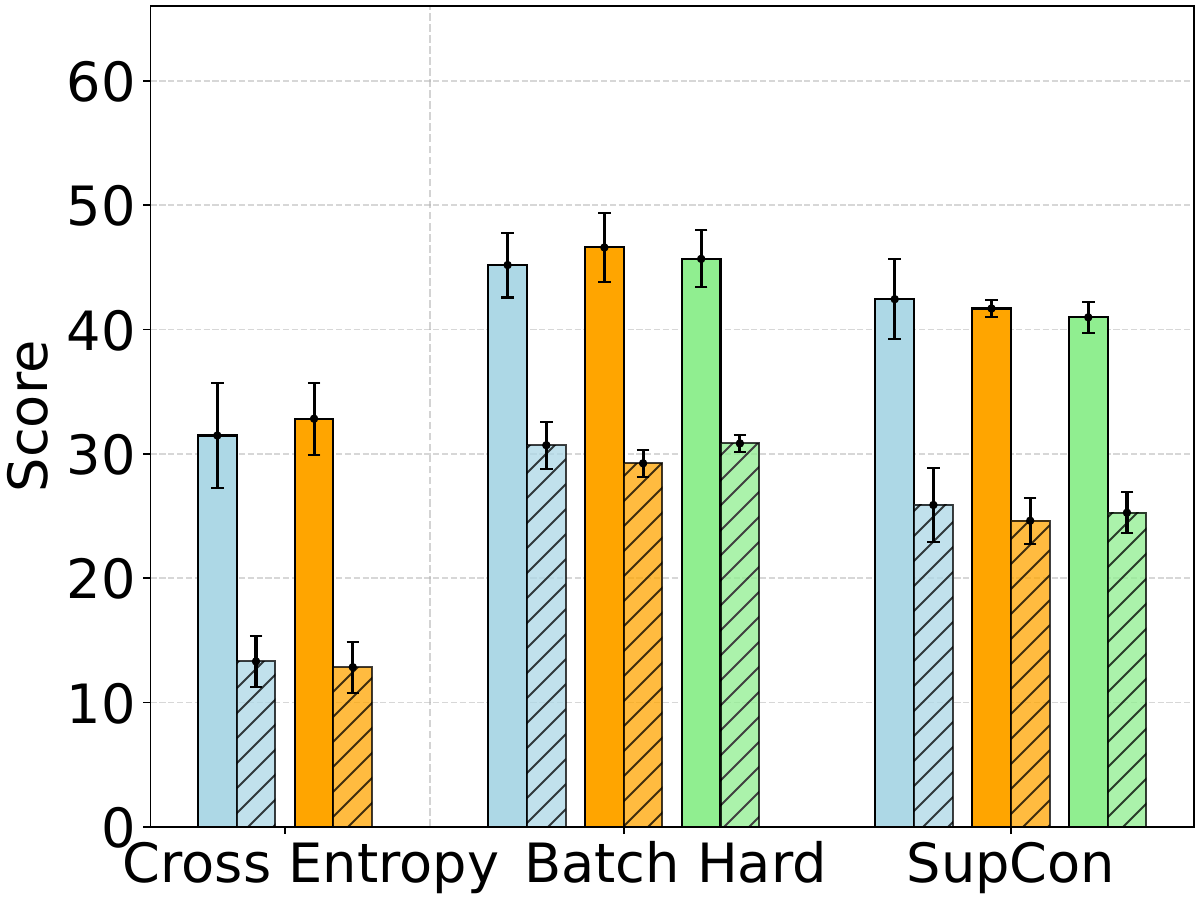}}
    \hfill
  \caption{Performance of recognizing entirely unseen states. Contrastive losses outperform the cross-entropy loss for all settings, and the ResNet backbone demonstrates significantly better generalization than the ViT.}
  \label{fig:generalization}
  \vspace{-0.45cm}
\end{figure}

\begin{figure*}
  \centering
  \vspace{5pt}
  \subfloat{%
       \includegraphics[width=0.99\linewidth]{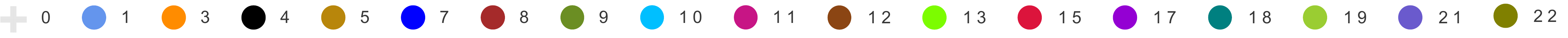}}
       \vspace{-0.2cm}
    \hfill
  \setcounter{subfigure}{0}
  \subfloat[\textbf{Trained with classification loss (cross-entropy)}. Intermediate states are grouped into once cluster, together with a significant amount of pre-defined assembly states. Furthermore, there is no path of intermediate states through the embedding space for the transition between pre-defined classes.
  \label{fig:umap_ce_nointer}]{%
        \includegraphics[width=0.49\linewidth, trim={0 0cm 0 0}]{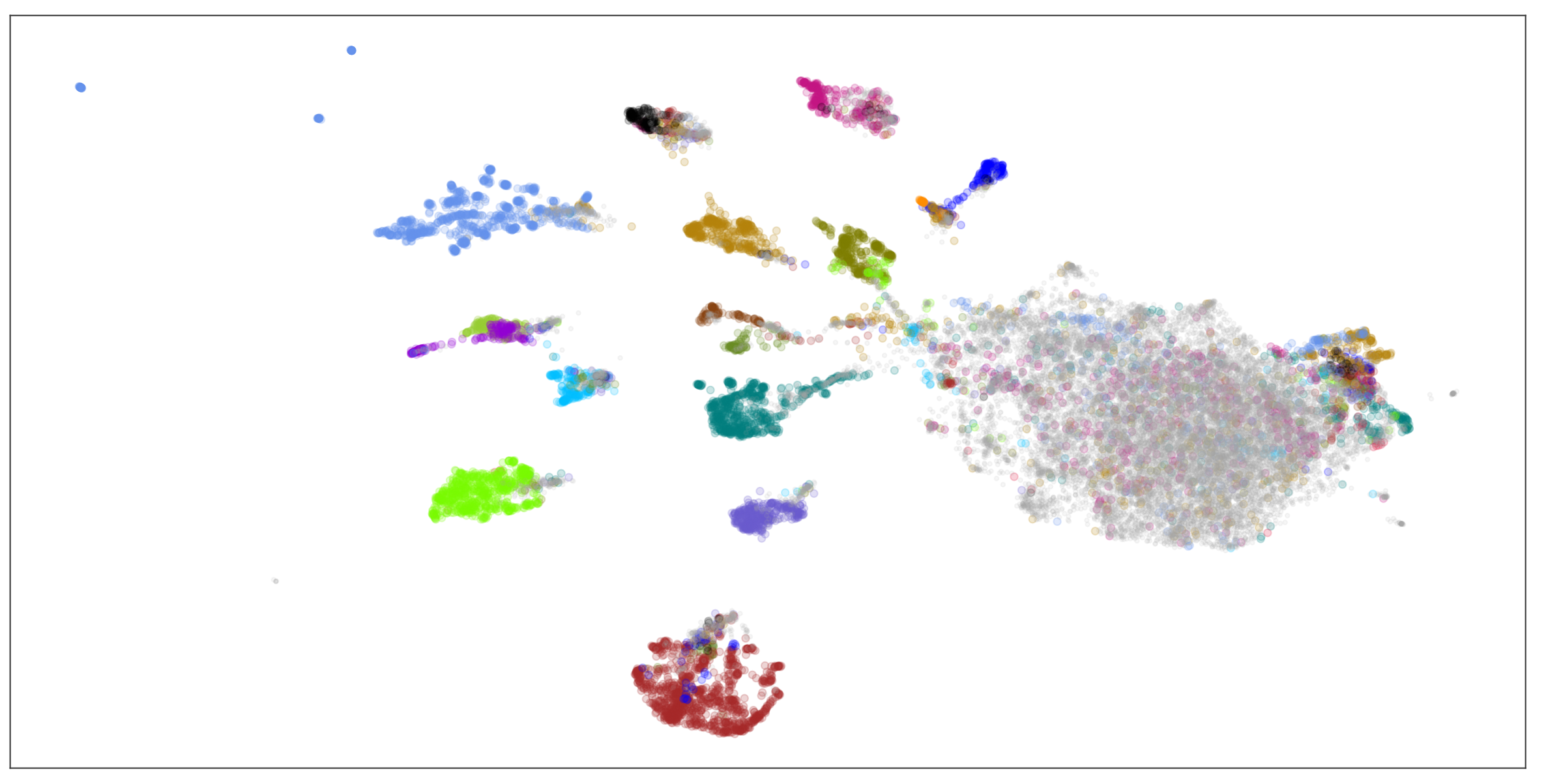}}
    \hfill
  \subfloat[\textbf{Trained with SupCon loss~\cite{supcon} without intermediate assembly states}. The embedding space shows paths between the pre-defined states, but poorly discriminates intermediate from pre-defined assembly states, with intermediates frequently located within defined classes. \label{fig:umap_nointer}]{%
        \includegraphics[width=0.49\linewidth, trim={0 0cm 0 0}]{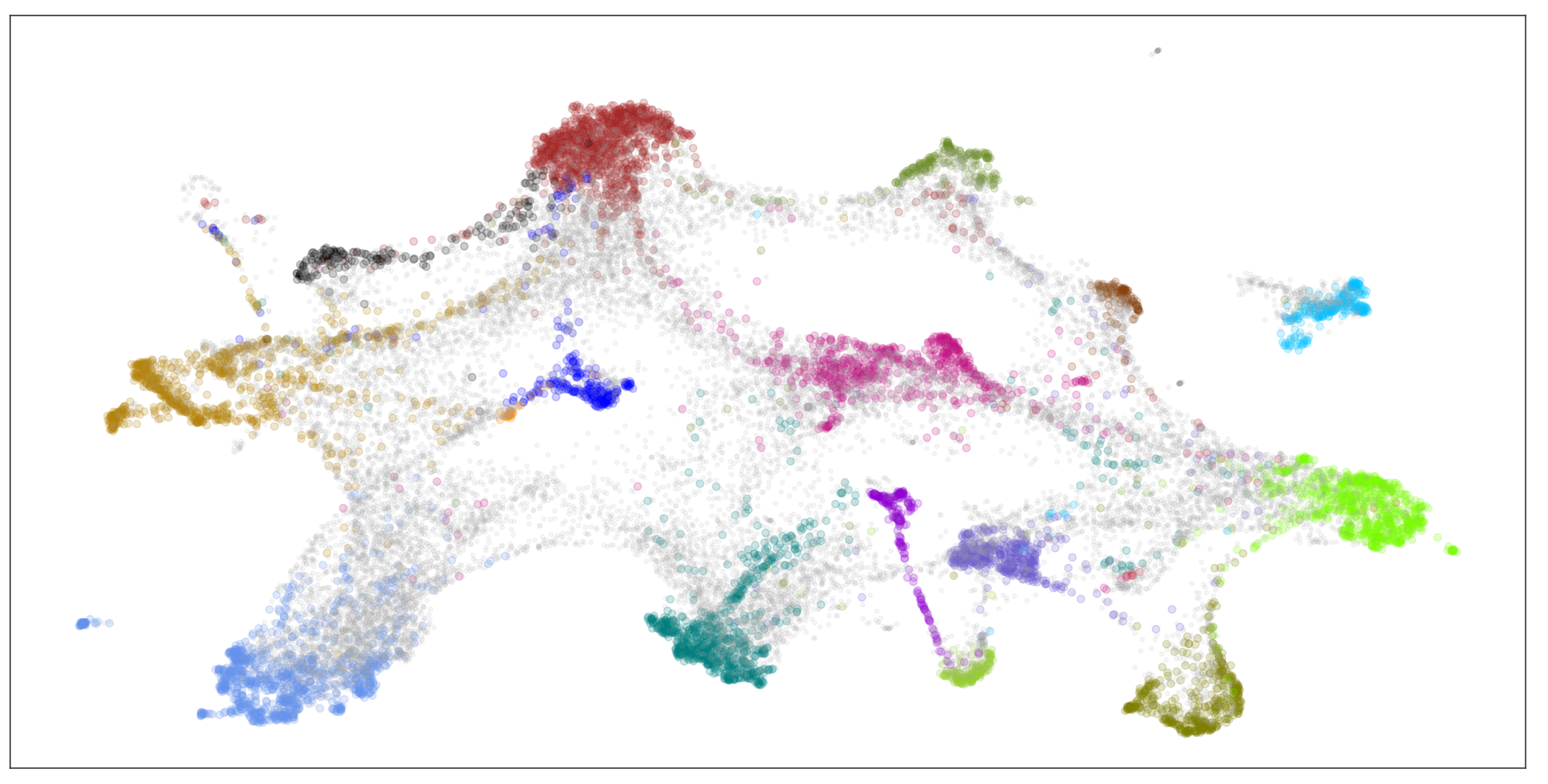}}
    \vspace{-0.2cm}
    \hfill
  \subfloat[\textbf{Trained with SupCon~\cite{supcon}, clustering all intermediate states together}. The model is able to distinguish between intermediate states and pre-defined clusters, but shows significantly reduced cluster quality for defined states.
 \label{fig:umap_inter}]{%
        \includegraphics[width=0.49\linewidth, trim={0 0.2cm 0 0.0cm}]{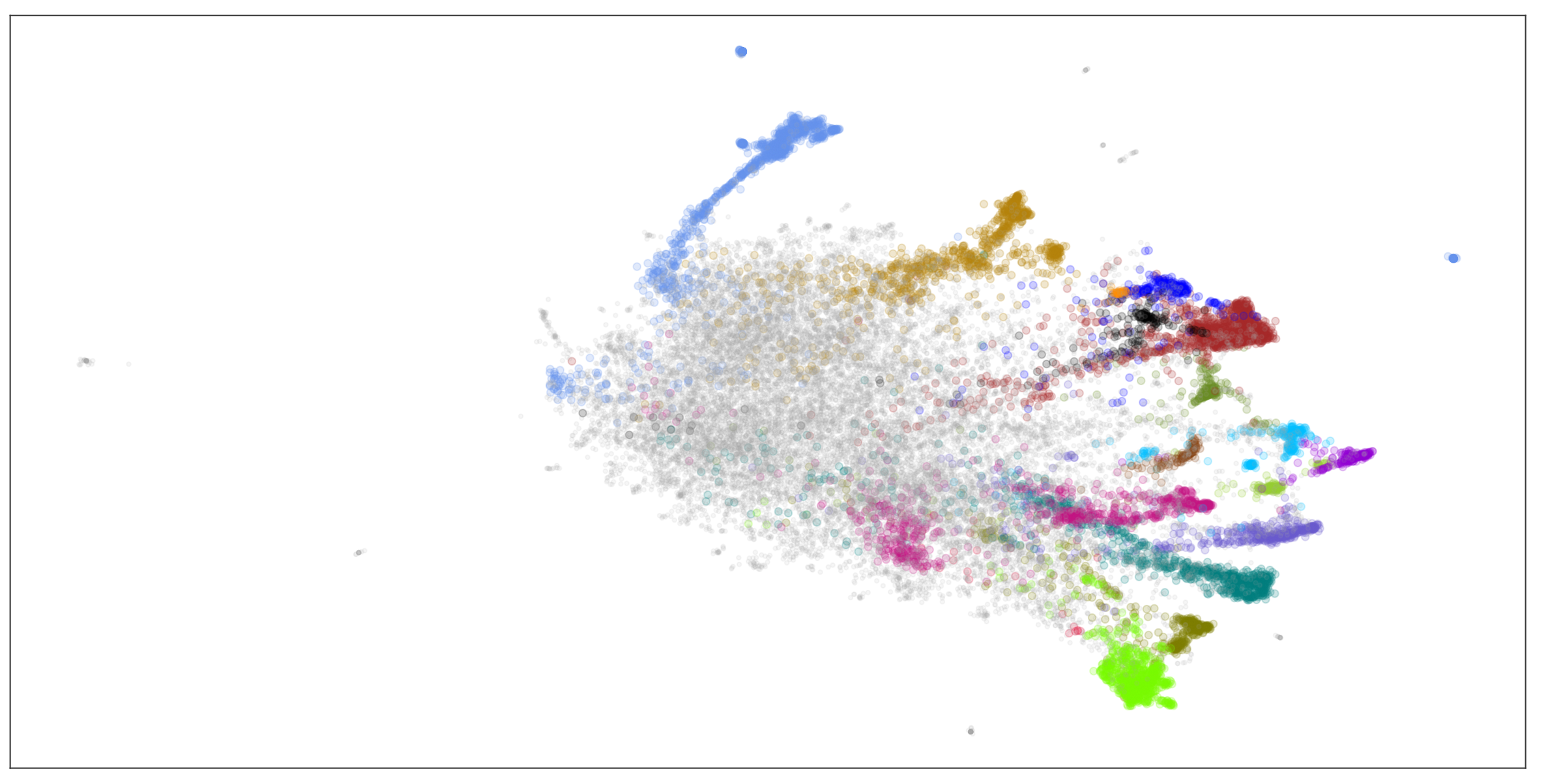}}
    \hfill
  \subfloat[\textbf{Trained with the SupCon loss~\cite{supcon} and the ISIL modification}. The modification retains the ability to discriminate between intermediate and defined states, whilst maintaining paths of undefined states through embedding space between pre-defined states.\label{fig:umap_interex}]{%
        \includegraphics[width=0.49\linewidth, trim={0 0.2cm 0 0.0cm}]{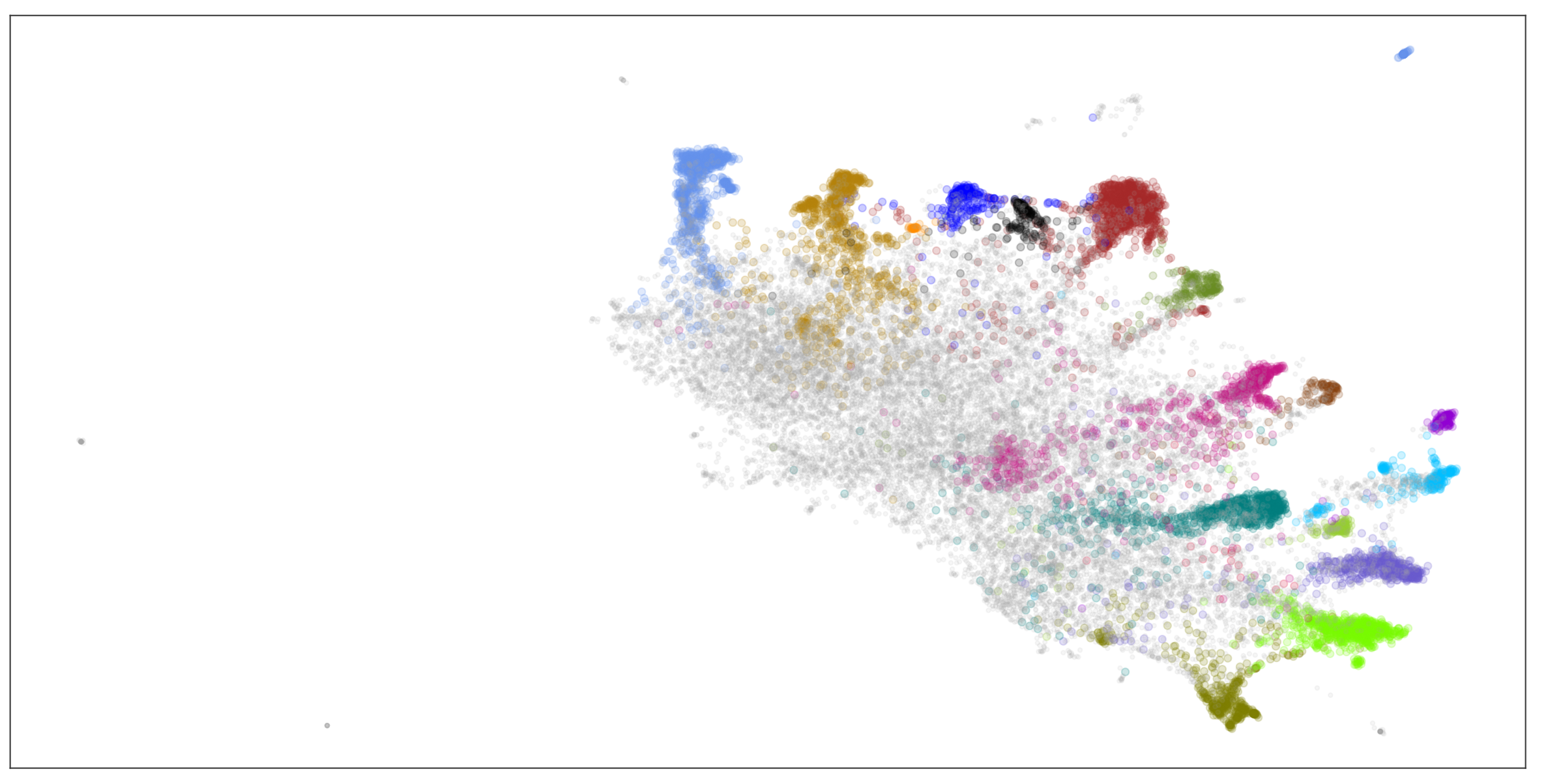}}
  \caption{UMAP~\cite{umap} visualization of the embedding space created by the best performing ViT backbone for various configurations. Intermediate states are indicated by the gray crosses, and some data points from defined states may appear as gray circles due to opacity.}
  \label{fig:umap_visualizations}
  \vspace{-0.5cm}
\end{figure*}

Additionally, regardless of the backbone type or loss, the clustering performance is improved by 5--22\% $\textit{MAP}\text{@}R(+)$ when the novel ISIL modification is applied. For $F_1\text{@}1$, the ISIL modification does not significantly affect performance for the ViT. For both backbones, the SupCon loss with ISIL performs best, clearly demonstrating the effectiveness of using undefined part configurations as negative samples.

The quantitative results described above can be further explained by comparing the embedding space generated by the models, shown in Fig.~\ref{fig:umap_visualizations}, by applying dimensionality reduction using unsupervised UMAP~\cite{umap}. The figure clearly demonstrates that discarding intermediate states during training, results in a large overlap of these states with defined assembly states (false positives). Clustering all intermediate states to a single group reduces the clustering quality of the remaining clusters, and results in a less meaningful embedding space. By applying the proposed ISIL modification, both the ability to discriminate between intermediate and pre-defined states, and a meaningful embedding space for intermediate representations, are retained.

\subsection{Generalization to Unseen States}
\label{sec:exp_generalization}
In this experiment, we test the hypothesis regarding improved generalization to entirely unseen part configurations for representation learning compared to classification-based approaches. Synthetic data is highly suitable for testing this generalization, since it provides full control over the data creation process. Specifically, we generate 18~new 3D~models of assembly states and use the same data generation methodology as IndustReal~\cite{schoonbeek2024industreal,borkman2021unity} to create 120~synthetic images per state, of which 100~are used for the reference set and 20~are kept for the query. The unseen states are modified versions of the IndustReal assembly states, \textit{e.g.}, one with screws placed differently, or a transposed position of an axle. 

The outcomes of this test are outlined in Fig.~\ref{fig:generalization} and clearly confirm the hypothesis. The performance increase for contrastive approaches range from 21--53\% on the classification metric $F_1\text{@}1$, with lower end of the range for ViT backbones. The increase in clustering performance ranges from a minimum of 85\% to a maximum of 204\% increased $\textit{MAP}\text{@}R$. The ResNet backbones outperform the ViT on generalization to new assembly configurations. These significant performance gains demonstrate the benefits of representation learning for assembly state recognition, and enables systems to generalize to new states (for instance, because of an updated procedure) without requiring the entire network to be re-trained.

\subsection{Generalization to Unseen Error States}
\label{sec:error_performance}
\begin{figure}
    \centering
    \vspace{7pt}
    \includegraphics[width=0.99\columnwidth, trim={0 6.5cm 0 0}]{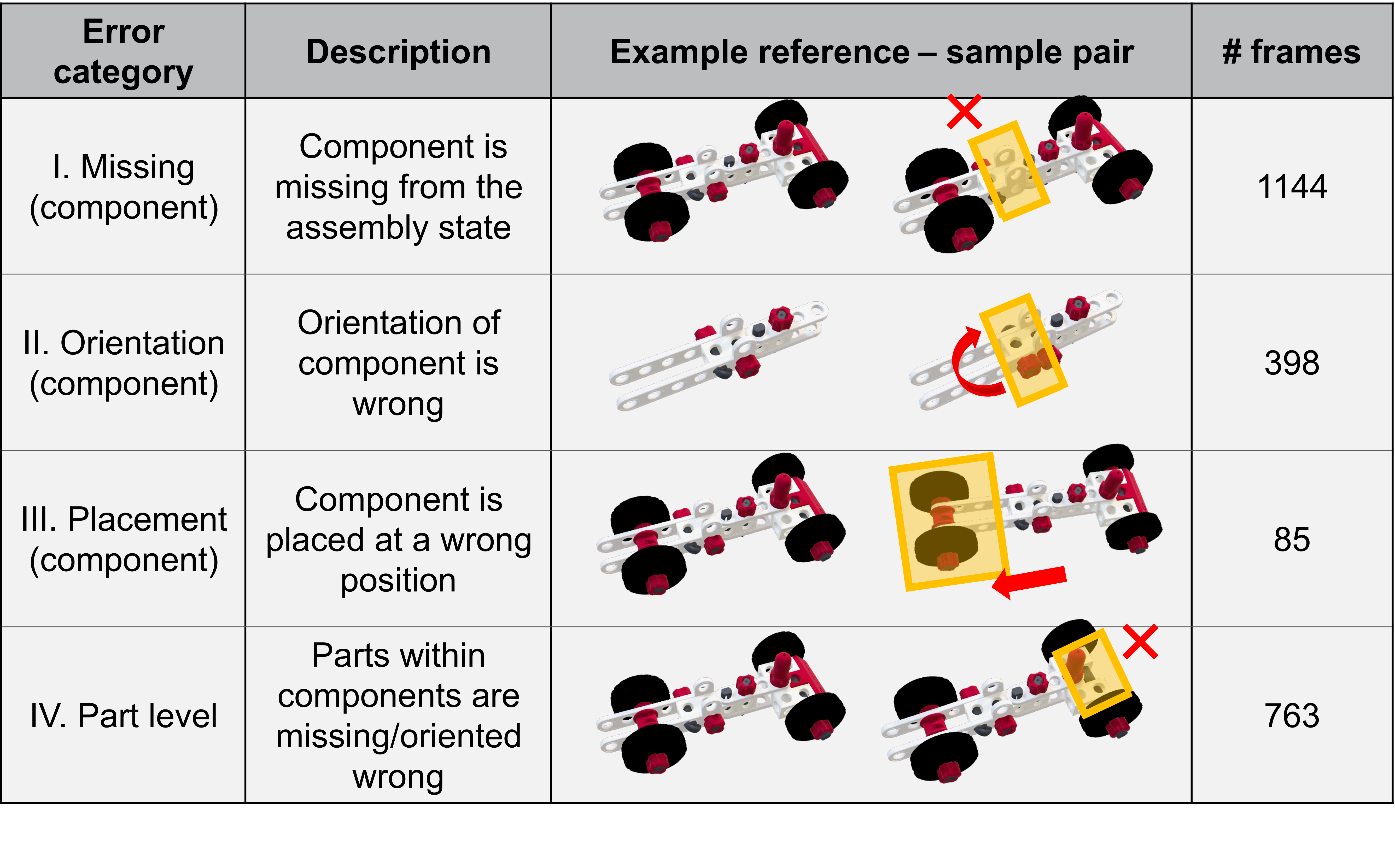}
    \caption{Categorization of errors including visual examples and the number of frames in IndustReal~\cite{schoonbeek2024industreal} for each category. Based on whether an error is at component or part level, they are classified into four different categories.}
    \vspace{-0.5cm}
    \label{fig:error_types}
\end{figure}

The performance of the proposed framework on unseen states provides a strong hint that distinction between erroneous and correct assembly states is feasible using this approach. After all, an execution error in an assembly procedure results in a new, unseen part configuration. To perform this experiment, at least two human annotators labeled the \textit{user-intended state} and \textit{error category} for every image in IndustReal~\cite{schoonbeek2024industreal} containing execution errors.
The former is annotated based on the work instructions for each video, and the latter based on four error categories: missing component~(\rom{1}.), incorrect orientation~(\rom{2}.), incorrect placement~(\rom{3}.), and part-level errors~(\rom{4}.). A description of the error categories and their occurrences in the IndustReal dataset are given in Fig.~\ref{fig:error_types}. The annotations are released with this work, facilitating future research on this topic.

The generalization of our framework to assembly errors is posed as a binary verification problem, where the cosine or Euclidian similarity between two correct states should be higher than the similarity between a correct and erroneous state. Our representation framework is used without any further fine-tuning, and the performance is quantified using the area under the precision-recall curve (AP). A single, error-free synthetic image is generated for every assembly state, with fixed lighting conditions and camera and object orientations. 
These images serve as \textit{anchor} images. 
Each image with an execution error in IndustReal~\cite{schoonbeek2024industreal} is a \textit{negative}, with the generated (error-free) image of the user-intended state as corresponding anchor to create the anchor-negative pair.
For each negative, exactly one correctly assembled, real-world image of the user-intended state is randomly sampled to serve as \textit{positive}, so that an equal number of positive and negative (error) pairs is sampled.
In this way, the dataset becomes balanced with 50\% anchor-positive and 50\% anchor-negative pairs, thereby preventing bias for models tending towards predicting non-corresponding states. A randomly performing model thus achieves an AP of~50.
Because the synthetic anchor images are always known to be defect-free, a model with an AP of 100 always has a higher similarity between the anchor and the error-free assembly state (positive) than between the anchor and the erroneous state (negative).

\begin{figure}
  \centering
  \vspace{7pt}
  \subfloat{%
       \includegraphics[width=0.99\linewidth, trim={0 0 0 0.0cm}]{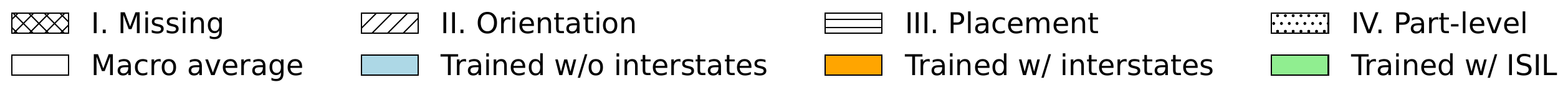}}
       \vspace{0.2cm}
    \hfill
  \setcounter{subfigure}{0}
  \subfloat{%
        \includegraphics[width=0.99\linewidth, trim={0 0.0cm 0 0.8cm}]{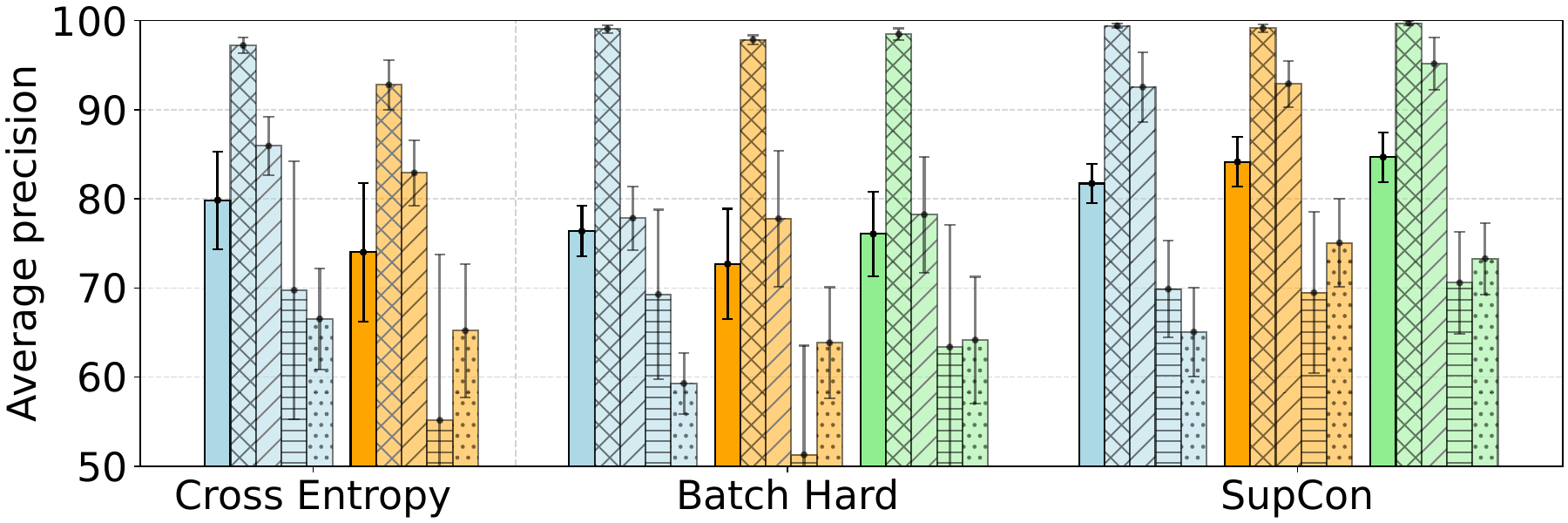}}
        \vspace{-0.1cm}
    \hfill
  \subfloat{%
        \includegraphics[width=0.99\linewidth, trim={0 0.6cm 0 0.8cm}]{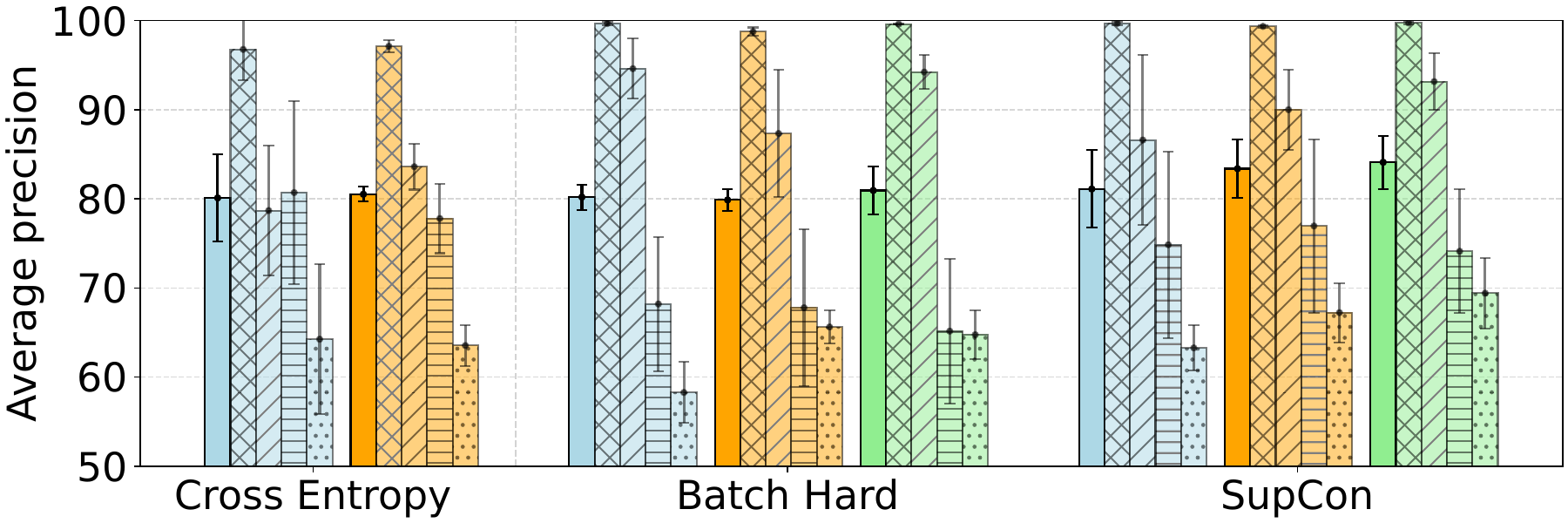}}
       \vspace{-0.1cm}
    \hfill
    \vspace{-0.45cm}
  \caption{Binary verification performance of ResNet (top) and ViT-S (bottom) between an error-free anchor image and a correct (positive) or erroneous assembly state (negative). The best contrastive models outperform the classification-based models and are trained with the ISIL modification.}
  \label{fig:error_performance}
  \vspace{-0.5cm}
\end{figure}

Figure~\ref{fig:error_performance} outlines the results for this experiment, where AP is reported for each error category, as defined in Fig.~\ref{fig:error_types}. The mean average precision for the best contrastive-trained ResNet is 6\% higher than that of the best ResNet trained for classification. Similarly, for the ViT, the best contrastive-trained model score increased 5\% compared to the best classification-trained model, reinforcing the benefits of representation learning with regards to assembly state recognition. The best AP for the ViT backbone is 0.58 lower than that of the best ResNet backbone. The ISIL modification improves the performance of both contrastive losses for both backbones. The results demonstrate that the proposed framework is able to accurately distinguish between correct assemblies and states with missing, incorrectly placed, and incorrectly oriented components. Furthermore, the results highlight the importance of investigating a models performance on erroneous assembly states.

\section{Discussion}
\label{sec:discussion}

In this work, we have addressed several challenges in assembly state recognition by approaching it as representation learning, rather than a classification problem. Furthermore, we have proposed the intermediate-state informed loss, a loss function modification that can be readily incorporated in existing contrastive loss functions. Our approach significantly outperforms classification-based approaches and generalizes to unseen part configurations. 
Additionally, the performance on error states is quantitatively evaluated with new error-state labels, demonstrating that the approach is able to distinguish between correct and erroneous states for various error categories, even when comparing real-world to synthetic images. The performance on the fine-grained part-level error category is lowest, but can likely be improved in future work by increasing the variety of assembly states during training, which can be readily generated for the synthetic training data. To stimulate such research, the new error-state labels are published together with this work.

The proposed ISIL modification demonstrates outstanding performance on the IndustReal test set and scores highest for both backbones on the error generalizations. Nonetheless, the observed benefits of ISIL on unseen synthetic assembly states are underwhelming. Grouping unlabeled states to a single cluster in feature space is expected to reduce the generalization of models, because new assembly states likely fall closely together in this cluster. Therefore, the ISIL modification is expected to show increased performance for generalization compared to models trained with intermediate states, but without ISIL. However, the results do not indicate a significant difference in performance, positive or negative, between discarding and using intermediate states, regardless of the ISIL modification. This observation is likely due to the absence of intermediate states in the generalization dataset. In future work, training on a subset of the states in IndustReal~\cite{schoonbeek2024industreal}, and using the rest of the data (including intermediate states) to test generalization, can verify this. Such an experiment can also confirm that the ISIL modification indeed creates a more meaningful embedding space.

\section{Conclusion}
\label{sec:conclusion}
Assembly state recognition has the potential to provide real-time assistance to numerous assembly and maintenance tasks, but requires substantial training data to learn the subtle differences between states. We propose to use representation learning towards solving this problem, and introduce the intermediate-state informed loss function modification (ISIL), effectively leveraging unlabeled transitions between assembly states by using these transitions as negative samples during training. The proposed framework outperforms classification-based approaches and generalizes to unseen states. Moreover, the performance on various assembly error types is investigated thoroughly.

In future work, it would be highly interesting to apply the proposed framework on a dataset containing real-world industrial procedures. Particularly, exploring procedures where the assembly consist of components with uniform texture and appearance, such as metal or plastic parts, would provide valuable insights. Additionally, we aim to provide additional weak supervision to the training by leveraging the timestamp of a sample in the procedure, as well as the (already annotated) previous and upcoming defined assembly states. With such modifications, we conjecture that a representation-learning based approach can effectively be integrated into edge devices to improve efficiency and significantly reduce chances of errors during industrial procedures, thereby providing meaningful assistance to workers.

\addtolength{\textheight}{-12cm}   

\section*{ACKNOWLEDGMENT}
The authors express their gratitude to Dan Lehman and Giacomo D'Amicantonio for their valuable insights. This work is partially executed ASML Research, with funding from ASML and TKI grant number TKI2112P07.

\bibliographystyle{IEEEtran}
\bibliography{IEEEabrv,main}

\begin{thebibliography}{10}
\providecommand{\url}[1]{#1}
\csname url@samestyle\endcsname
\providecommand{\newblock}{\relax}
\providecommand{\bibinfo}[2]{#2}
\providecommand{\BIBentrySTDinterwordspacing}{\spaceskip=0pt\relax}
\providecommand{\BIBentryALTinterwordstretchfactor}{4}
\providecommand{\BIBentryALTinterwordspacing}{\spaceskip=\fontdimen2\font plus
\BIBentryALTinterwordstretchfactor\fontdimen3\font minus \fontdimen4\font\relax}
\providecommand{\BIBforeignlanguage}[2]{{%
\expandafter\ifx\csname l@#1\endcsname\relax
\typeout{** WARNING: IEEEtran.bst: No hyphenation pattern has been}%
\typeout{** loaded for the language `#1'. Using the pattern for}%
\typeout{** the default language instead.}%
\else
\language=\csname l@#1\endcsname
\fi
#2}}
\providecommand{\BIBdecl}{\relax}
\BIBdecl

\bibitem{su2019deep}
Y.~Su, J.~Rambach, N.~Minaskan, P.~Lesur, A.~Pagani, and D.~Stricker, ``Deep multi-state object pose estimation for augmented reality assembly,'' in \emph{2019 IEEE Int. Symp. on Mixed and Augmented Reality Adjunct (ISMAR-Adjunct)}.\hskip 1em plus 0.5em minus 0.4em\relax IEEE, 2019, pp. 222--227.

\bibitem{stanescu2023state}
A.~Stanescu, P.~Mohr, M.~Kozinski, S.~Mori, D.~Schmalstieg, and D.~Kalkofen, ``State-aware configuration detection for augmented reality step-by-step tutorials,'' in \emph{2023 IEEE Int. Symp. on Mixed and Augmented Reality (ISMAR)}.\hskip 1em plus 0.5em minus 0.4em\relax IEEE, 2023, pp. 157--166.

\bibitem{liu2020tga}
H.~Liu, Y.~Su, J.~Rambach, A.~Pagani, and D.~Stricker, ``Tga: Two-level group attention for assembly state detection,'' in \emph{2020 IEEE Int. Symp. on Mixed and Augmented Reality Adjunct (ISMAR-Adjunct)}.\hskip 1em plus 0.5em minus 0.4em\relax IEEE, 2020, pp. 258--263.

\bibitem{zhou2020fine}
B.~Zhou and S.~G{\"u}ven, ``Fine-grained visual recognition in mobile augmented reality for technical support,'' \emph{{IEEE} Trans. Vis. Comput. Graphics}, vol.~26, no.~12, pp. 3514--3523, 2020.

\bibitem{pang2020marker}
J.~Pang, J.~Zhang, Y.~Li, and W.~Sun, ``A marker-less assembly stage recognition method based on segmented projection contour,'' \emph{Advanced Engineering Informatics}, vol.~46, p. 101149, 2020.

\bibitem{murray2024equipment}
K.~Murray, J.~Schierl, K.~Foley, and Z.~Duric, ``Equipment assembly recognition for augmented reality guidance,'' in \emph{2024 IEEE Int. Conf. on Artificial Intelligence and eXtended and Virtual Reality (AIxVR)}.\hskip 1em plus 0.5em minus 0.4em\relax IEEE, 2024, pp. 109--118.

\bibitem{jones2021fine}
J.~D. Jones, C.~Cortesa, A.~Shelton, B.~Landau, S.~Khudanpur, and G.~D. Hager, ``Fine-grained activity recognition for assembly videos,'' \emph{IEEE Robotics and Automation Letters}, vol.~6, no.~2, pp. 3728--3735, 2021.

\bibitem{musgrave2020metric}
K.~Musgrave, S.~Belongie, and S.-N. Lim, ``A metric learning reality check,'' in \emph{Computer Vision--ECCV 2020: 16th European Conf., Glasgow, UK, August 23--28, 2020, Proceedings, Part XXV 16}.\hskip 1em plus 0.5em minus 0.4em\relax Springer, 2020, pp. 681--699.

\bibitem{schoonbeek2024industreal}
T.~J. Schoonbeek, T.~Houben, H.~Onvlee, F.~van~der Sommen \emph{et~al.}, ``Industreal: A dataset for procedure step recognition handling execution errors in egocentric videos in an industrial-like setting,'' in \emph{Proc. of the IEEE/CVF Winter Conf. on Applications of Computer Vision}, 2024, pp. 4365--4374.

\bibitem{schieber2024asdf}
H.~Schieber, S.~Li, N.~Corell, P.~Beckerle, J.~Kreimeier, and D.~Roth, ``Asdf: Assembly state detection utilizing late fusion by integrating 6d pose estimation,'' \emph{arXiv preprint arXiv:2403.16400}, 2024.

\bibitem{pang2021image}
J.~Pang, S.-K. Ong, and A.~Y.-C. Nee, ``Image and model sequences matching for on-site assembly stage identification,'' \emph{Robotics and Computer-Integrated Manufacturing}, vol.~72, p. 102185, 2021.

\bibitem{elhafsi2023semantic}
A.~Elhafsi, R.~Sinha, C.~Agia, E.~Schmerling, I.~A. Nesnas, and M.~Pavone, ``Semantic anomaly detection with large language models,'' \emph{Autonomous Robots}, vol.~47, no.~8, pp. 1035--1055, 2023.

\bibitem{simclr}
T.~Chen, S.~Kornblith, M.~Norouzi, and G.~Hinton, ``A simple framework for contrastive learning of visual representations,'' in \emph{Int. Conf. on machine learning}.\hskip 1em plus 0.5em minus 0.4em\relax PMLR, 2020, pp. 1597--1607.

\bibitem{simclrv2}
T.~Chen, S.~Kornblith, K.~Swersky, M.~Norouzi, and G.~E. Hinton, ``Big self-supervised models are strong semi-supervised learners,'' \emph{Adv. Neural Inf. Process. Syst.}, vol.~33, pp. 22\,243--22\,255, 2020.

\bibitem{dino}
M.~Caron, H.~Touvron, I.~Misra, H.~J{\'e}gou, J.~Mairal, P.~Bojanowski, and A.~Joulin, ``Emerging properties in self-supervised vision transformers,'' in \emph{Proc. of the IEEE/CVF Int. Conf. on computer vision}, 2021, pp. 9650--9660.

\bibitem{snn1}
J.~Bromley, I.~Guyon, Y.~LeCun, E.~S{\"a}ckinger, and R.~Shah, ``Signature verification using a" siamese" time delay neural network,'' \emph{Adv. Neural Inf. Process. Syst.}, vol.~6, 1993.

\bibitem{oneshotsiamese}
G.~Koch, R.~Zemel, R.~Salakhutdinov \emph{et~al.}, ``Siamese neural networks for one-shot image recognition,'' in \emph{ICML deep learning workshop}, vol.~2.\hskip 1em plus 0.5em minus 0.4em\relax Lille, 2015, p.~0.

\bibitem{snn2}
S.~Chopra, R.~Hadsell, and Y.~LeCun, ``Learning a similarity metric discriminatively, with application to face verification,'' in \emph{2005 IEEE Computer Society Conf. on Computer Vision and Pattern Recognition (CVPR'05)}, vol.~1.\hskip 1em plus 0.5em minus 0.4em\relax IEEE, 2005, pp. 539--546.

\bibitem{facenet}
F.~Schroff, D.~Kalenichenko, and J.~Philbin, ``Facenet: A unified embedding for face recognition and clustering,'' in \emph{Proc. of the IEEE Conf. on computer vision and pattern recognition}, 2015, pp. 815--823.

\bibitem{triplet}
A.~Hermans, L.~Beyer, and B.~Leibe, ``In defense of the triplet loss for person re-identification,'' \emph{arXiv preprint arXiv:1703.07737}, 2017.

\bibitem{infonce}
A.~v.~d. Oord, Y.~Li, and O.~Vinyals, ``Representation learning with contrastive predictive coding,'' \emph{arXiv preprint arXiv:1807.03748}, 2018.

\bibitem{supcon}
P.~Khosla, P.~Teterwak, C.~Wang, A.~Sarna, Y.~Tian, P.~Isola, A.~Maschinot, C.~Liu, and D.~Krishnan, ``Supervised contrastive learning,'' \emph{Adv. Neural Inf. Process. Syst.}, vol.~33, pp. 18\,661--18\,673, 2020.

\bibitem{borkman2021unity}
S.~Borkman, A.~Crespi, S.~Dhakad, S.~Ganguly, J.~Hogins, Y.-C. Jhang, M.~Kamalzadeh, B.~Li, S.~Leal, P.~Parisi \emph{et~al.}, ``Unity perception: Generate synthetic data for computer vision,'' \emph{arXiv preprint arXiv:2107.04259}, 2021.

\bibitem{he2016deep}
K.~He, X.~Zhang, S.~Ren, and J.~Sun, ``Deep residual learning for image recognition,'' in \emph{Proc. of the IEEE Conf. on computer vision and pattern recognition}, 2016, pp. 770--778.

\bibitem{dosovitskiy2020image}
A.~Dosovitskiy, L.~Beyer, A.~Kolesnikov, D.~Weissenborn, X.~Zhai, T.~Unterthiner, M.~Dehghani, M.~Minderer, G.~Heigold, S.~Gelly \emph{et~al.}, ``An image is worth 16x16 words: Transformers for image recognition at scale,'' \emph{arXiv preprint arXiv:2010.11929}, 2020.

\bibitem{vit}
A.~Steiner, A.~Kolesnikov, , X.~Zhai, R.~Wightman, J.~Uszkoreit, and L.~Beyer, ``How to train your vit? data, augmentation, and regularization in vision transformers,'' \emph{arXiv preprint arXiv:2106.10270}, 2021.

\bibitem{loshchilov2016sgdr}
I.~Loshchilov and F.~Hutter, ``Sgdr: Stochastic gradient descent with warm restarts,'' \emph{arXiv preprint arXiv:1608.03983}, 2016.

\bibitem{umap}
L.~McInnes, J.~Healy, and J.~Melville, ``Umap: Uniform manifold approximation and projection for dimension reduction,'' \emph{arXiv preprint arXiv:1802.03426}, 2018.

\end{thebibliography}

\end{document}